\documentclass{article}

\usepackage{arxiv}

\usepackage[utf8]{inputenc} 
\usepackage[T1]{fontenc}    
\usepackage{hyperref}       
\usepackage{url}            
\usepackage{booktabs}       
\usepackage{amsfonts}       
\usepackage{nicefrac}       
\usepackage{microtype}      
\usepackage{lipsum}		
\usepackage{graphicx}
\usepackage[numbers]{natbib}
\usepackage{doi}
\usepackage{enumitem}
\setlist{leftmargin=3.0mm}

\title{Size doesn't matter: predicting physico- or biochemical properties based on dozens of molecules}

\author{ \href{https://orcid.org/0000-0002-4442-5807}{\includegraphics[scale=0.06]{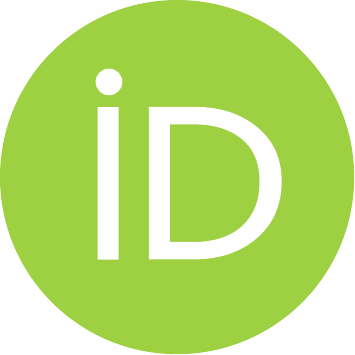}\hspace{1mm}Kirill Karpov}\thanks{\textit{Email address}: \texttt{karpov@radio.chem.msu.ru}} \\
	Lomonosov Moscow State University\\
	Department of Chemistry\\
	Leninskie gory, 1 bld. 3\\
	Moscow 119991, Russia\\
	\And
	\href{https://orcid.org/0000-0001-8891-6862}{\includegraphics[scale=0.06]{orcid.pdf}\hspace{1mm}Artem Mitrofanov} \\
	Lomonosov Moscow State University\\
	Department of Chemistry\\
	Leninskie gory, 1 bld. 3\\
	Moscow 119991, Russia\\
	\And
	\href{https://orcid.org/0000-0001-6117-5662}{\includegraphics[scale=0.06]{orcid.pdf}\hspace{1mm}Vadim Korolev} \\
	Lomonosov Moscow State University\\
	Department of Chemistry\\
	Leninskie gory, 1 bld. 3\\
	Moscow 119991, Russia\\
	\And
	\href{https://orcid.org/0000-0003-4265-235X}{\includegraphics[scale=0.06]{orcid.pdf}\hspace{1mm}Valery Tkachenko} \\
	Science Data Software, LLC\\
	14909 Forest Landing Cir\\
	Rockville, MD 20850, USA\\
}

\renewcommand{\shorttitle}{Size doesn't matter: predicting physico- or biochemical properties}

\hypersetup{
pdftitle={Size doesn't matter: predicting physico- or biochemical properties based on dozens of molecules},
pdfsubject={q-bio.NC, q-bio.QM},
pdfauthor={Kirill~Karpov, Artem~Mitrofanov, Vadim~Korolev, Valery~Tkachenko},
pdfkeywords={QSAR/QSPR, Graph-convolutional neural networks, Transfer leraning},
}

\begin{document}
\maketitle

\begin{abstract}
The use of machine learning in chemistry has become a common practice. At the same time, despite the success of modern machine learning methods, the lack of data limits their use. Using a transfer learning methodology can help solve this problem. This methodology assumes that a model built on a sufficient amount of data captures general features of the chemical compound structure on which it was trained and that the further reuse of these features on a dataset with a lack of data will greatly improve the quality of the new model. In this paper, we develop this approach for small organic molecules, implementing transfer learning with graph convolutional neural networks. The paper shows a significant improvement in the performance of models for target properties with a lack of data. The effects of the dataset composition on model’s quality and the applicability domain of the resulting models are also considered.
\end{abstract}


\section{Introduction}
Machine learning (ML) methods proved to be a useful tool in chemistry \cite{vamathevan2019applications,lipinski2019advances,gomez2018automatic,gao2018using} and materials science \cite{schmidt2019recent,liu2017materials,janet2020accurate}. Their use is mainly focused on determining the relation between the structure of a chemical compound and a certain property \cite{neves2018qsar,muratov2020qsar}. Further, found relations may be applied for searching new compounds with the desired value of the property. Recently, the artificial neural networks (ANN) has become one of the most popular  (ML) approaches due to the high accuracy and the flexibility of the architectures \cite{lecun2015deep,korotcov2017comparison}. The ANN structure is a set of layers, each of them consisting of one or more nodes, and each node, in turn, can be described by numerical values - weights and biases. These values are fitted during the training process to minimize the deviation between the target value and its approximation). The layered structure allows to increase the level of abstraction layer by layer going from the description of the object to its properties.	One of the main limitations for the application of ANN in the modern cheminformatics is a demand for  large sets of data (hundreds to millions data points) for model training \cite{muratov2020qsar,butler2018machine,yang2019concepts}. At the same time, most of the chemical and biological properties, do not have enough experimental data collected; the size of datasets for such properties often does not exceed tens or hundreds of compounds. This list includes substances related to the biological target activities, or the substances for which synthesis and measurement is difficult or impossible for some reasons \cite{andreadi2020heavy,marchenko2020database}. One of the most promising methods of working with such datasets is the transfer learning approach. Recently, this methodology has found major interest in the scientific community, including chemistry and materials science fields \cite{cai2020transfer,pesciullesi2020transfer,lentelink2020transfer}. Its main idea is to transfer part of the information obtained from a large dataset to a problem with the lack of training data. Generally, it is represented in the transfer of some parameters such as mentioned above weights and biases from a pre-trained model (‘donor’ model) to a model that is supposed to be trained on a small dataset (‘acceptor’ model). As it is known from the image recognition field, the procedure works well for similar datasets \cite{weiss2016survey}. In case of chemistry, it is supposed to transfer general information about chemical structures to the acceptor model, leaving algorithm to fit it for the specific property. It should be noted that in addition to the property similarity of the donor and acceptor dataset, it is necessary that the donor dataset should be as large and diverse as possible. This helps to generalize transferable parameters of the model and thus increases the applicability domain (AD) of the donor model, which will consequently improve the quality and the AD of the acceptor model.

The transfer learning approach for chemical tasks may be performed with different types of neural network architectures, from the simplest feed-forward ANN (FFANN) \cite{muegge2016overview,yamada2019predicting} to more domain-specific graph convolutional neural networks (GCNN) \cite{zhou2020graph} [Fig. 1 (b)]. GCNN may be applied to a wide range of objects that can be represented in graph form and as a consequence do not need a separate feature extraction algorithm to work with molecules. Another advantage of these models is the ability to hold object’s information within first graph convolutional layers that can be especially useful in transfer learning tasks. GCNN models have been tested for predicting properties of materials and small organic molecules \cite{korolev2019graph,lee2021transfer,xie2018crystal,chen2019graph,lu2020coupling,devogelaer2020co} and also showed promising results for transfer learning applications in materials science.

\begin{figure}[ht]
  \centering
  \includegraphics[height=10cm]{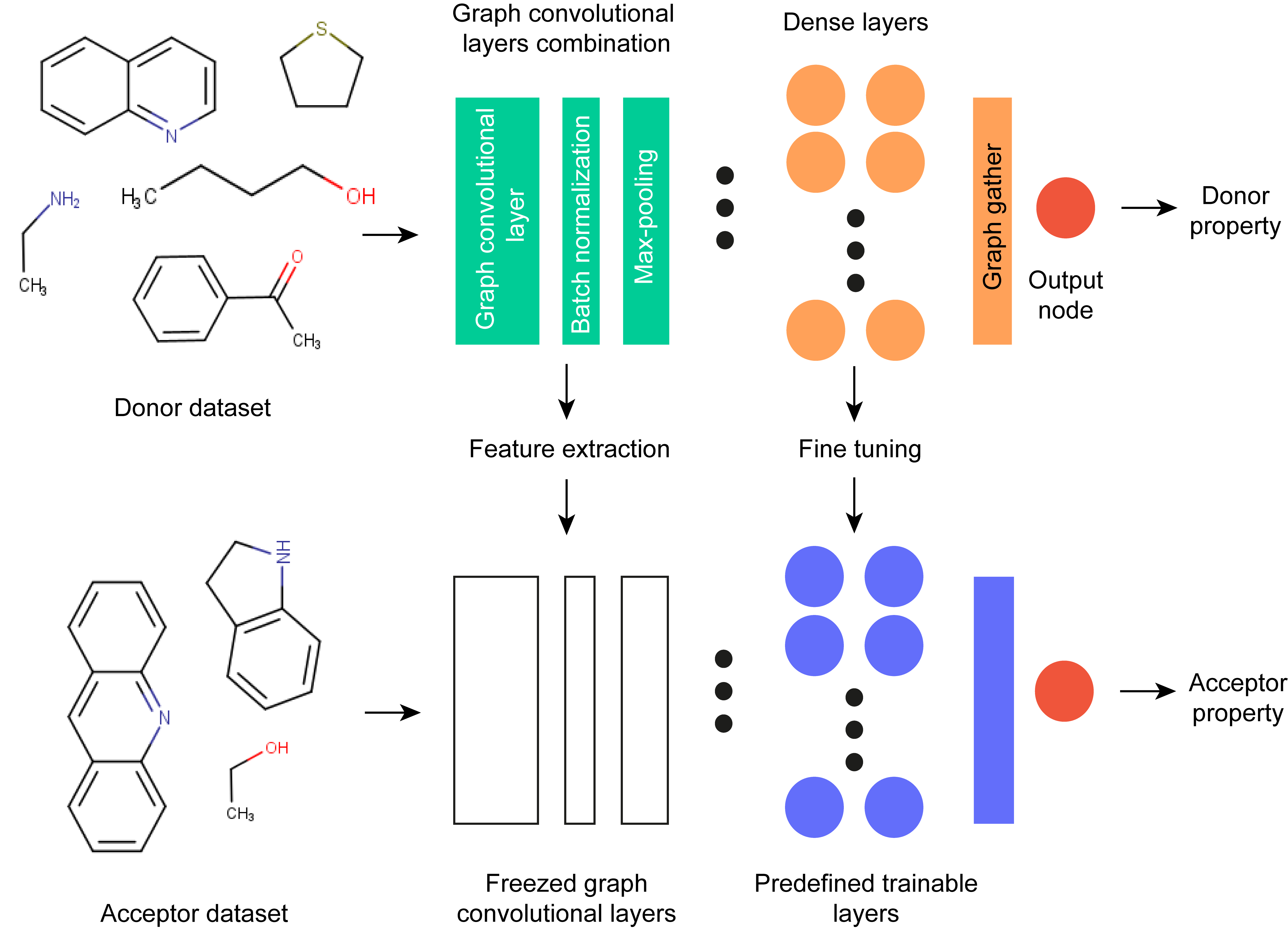}
  \caption{General workflow.}
  \label{fig:fig1}
\end{figure}

In this study, we would like to expand the use of GCNN architecture with transfer learning technique for small organic molecules and determine the influence of the datasets’ size, diversity and similarity on the final accuracy and AD. For this purpose, we trained set of transfer models on various physicochemical and biomedical properties in combination with different training set sizes and layers compositions.

\section{Methods}
Transfer learning approach may be generally represented by two techniques – feature extraction and fine tuning \cite{yosinski2014transferable}. Feature extraction is the procedure in which parameters of several donor models’ layers are frozen during training on acceptor dataset, which causes the layers to behave like feature extraction algorithm for target objects \cite{li2020inductive,lukas2019deep}. Fine tuning donor model generally retrains acceptor dataset as training set and models’ parameters obtained during previous training usually act as initial. In this work we used both techniques.

Standard model structure presented in Fig.1 this network architecture was inspired by \cite{korolev2019graph}. All additional training scripts were written in Python, using deepchem \cite{Ramsundar-et-al-2019}  and tensorflow \cite{tensorflow2015-whitepaper} frameworks, code presented in https://github.com/SmartChemDesign/gcnn\_mol\_transfer. The model quality was estimated with coefficient of determination ($R^2$ score) for regressions and area under receiver operating characteristic curve (ROC AUC score) for classifications.

The transfer learning performance was tested on a set of common physicochemical and biomedical properties of small organic molecules. We selected several popular datasets marked up both for classification and regression tasks. In particular, regression datasets are presented by OPERA PHYSPROP database \cite{mansouri2018opera}, that contains experimental values for the following properties: water solubility, bioconcentration factor, biodegradability half-life, Henry’s Law constant, fish biotransformation half-life, octanol–air partition coefficient, soil adsorption coefficient and vapor pressure.  Biochemical classification data was obtained from PubChem bioassays datasets \cite{kim2021pubchem}. It included data on activities versus: 1851(2c19) - Cytochrome P450, family 2, subfamily C, polypeptide 19, 1851(2c9) - Cytochrome P450, family 2, subfamily C, polypeptide 9, 1851(3a4) - Cytochrome P450, family 3, subfamily A, polypeptide 4, 1851(1a2) - Cytochrome P450, family 1, subfamily A, polypeptide 2, 1851(2d6) - Cytochrome P450, family 2, subfamily D, polypeptide 6, isoform 2. The bioassays datasets were selected due to their endpoint’s similarity, which makes it possible to use one of them as a donor for the others. We used 1.5 million structures from ChEMBL \cite{davies2015chembl, gaulton2012chembl} with the known values of octanol-water partition coefficient (logP)  as a donor dataset for OPERA database and 1851(1a2) for bioassays datasets.

To demonstrate and analyze the operation of the algorithm, the donor datasets were split in the same manner. We extracted 4 subsets of 10, 20, 50 and 100 molecules from every dataset with the highest diversity of the target property (we also tested other subsampling approaches as discussed below). Obtained small subsets were used as training sets while the rest of the data were used as an external test set. In all the cases, the published metrics refer to the test sets. 

\begin{figure}[ht]
  \centering
  \includegraphics[height=6cm]{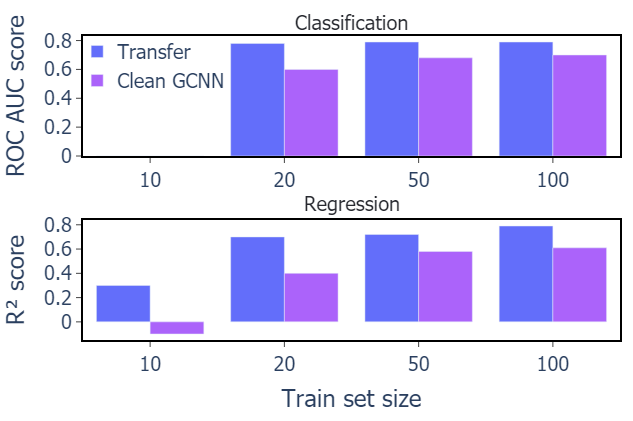}
  \caption{Comparison between transfer learning model and pure GCNN model performance.}
  \label{fig:fig2}
\end{figure}

\section{Results and discussions}
At the preliminary stage, we would like to verify the general possibility to improve the model’s quality with the transfer learning technique, as well as to find the best architecture of the neural network (the number of nodes and layers). We used data on water solubility (logS) as acceptor dataset. This dataset should be quite similar to the donor one since the acceptor model quality should depend on endpoints similarity. We also could use druglikeness \cite{ursu2011understanding} concept to transform continuous to categorical values for both logP and logS which is necessary for training classification models. Thus, we trained logP classification and regression models. 

\begin{figure}[ht]
  \centering
  \includegraphics[height=6cm]{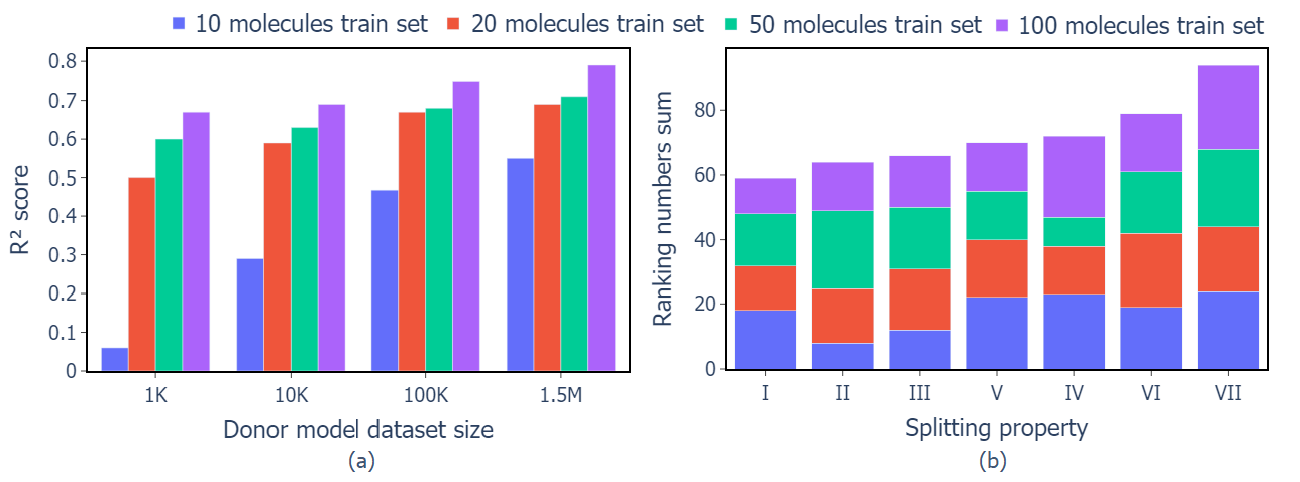}
  \caption{(a) Acceptor models quality depending on donor dataset size; (b) Sum of places of transfer learning models depending on splitting property (I – aromatic rings number, II – molecular weight, III - rotatable bonds number, IV - hydrogen-bond acceptor number, V - heterocycles number, VI - potential surface area, VII - hydrogen-bond donor number) for different train set sizes.}
  \label{fig:fig3}
\end{figure}

We used a GCNN with 2 graph convolutional layers followed by 3 dense layers. We also took a series of solubility datasets split as mentioned in the Methods section. Based on the resulting partition, we built two series of models, with and without transfer learning. The results are presented in Fig. 2.
In all the cases, the transfer learning approach improved the quality of the models, proving the general concept. But the simple tests with the random subsampling of data for acceptor models were not very indicative in a quantitative meaning, due to possibility of choosing similar molecules for training. 

In order to eliminate uncertainty in the similarity of the donor and acceptor datasets compositions, several additional tests were carried out. First of all, the influence of the donor dataset on transfer learning was studied. To determine how the size of the donor dataset affects the quality of the acceptor model, a series of models was built based on the logP dataset with different numbers of molecules randomly taken from the main database. As a result, 4 models were obtained, built on different samples of 1,500,000, 100,000, 10,000, and 1,000 molecules, respectively. Previously used logS datasets with a training set size of 10, 20, 50, and 100 molecules were taken as acceptor datasets. For each acceptor dataset, 4 acceptor models were built for each of the obtained donors. The results are presented in Fig. 3 (a). The figure shows a steady trend of improving the quality of the acceptor model with an increase in the number of molecules in the donor dataset, which confirms our assumptions. Moreover, this dependence is most strongly visible for the smallest acceptor datasets. 
At the previous stages we used the random subsampling of the acceptor data based on the endpoint diversity. But the applied problems rarely provide an opportunity to choose a convenient dataset. Thus, we studied how the composition of the acceptor dataset affects model quality, to provide an opportunity to assess the suitability of the available small data for transfer learning. First of all, the following set of splitting properties were selected: molecular weight, number of aromatic rings, hydrogen bond acceptor number, hydrogen bond donor number, number of heterocycles, number of rotatable bonds, potential surface area. The choice of the splitting markers was based both on the diversity of the properties and the simplicity of their calculation. Then we split acceptor datasets in such a way that training subset contains the most diverse molecules according to the selected properties, one property for every partition. Benchmark was carried out for the ChEMBL logP and 1851(1a2) as donor datasets and the whole list of OPERA with the rest of bioassays as acceptors. The results are presented in Fig. 3 (b).

\begin{figure}[ht]
  \centering
  \includegraphics[height=6cm]{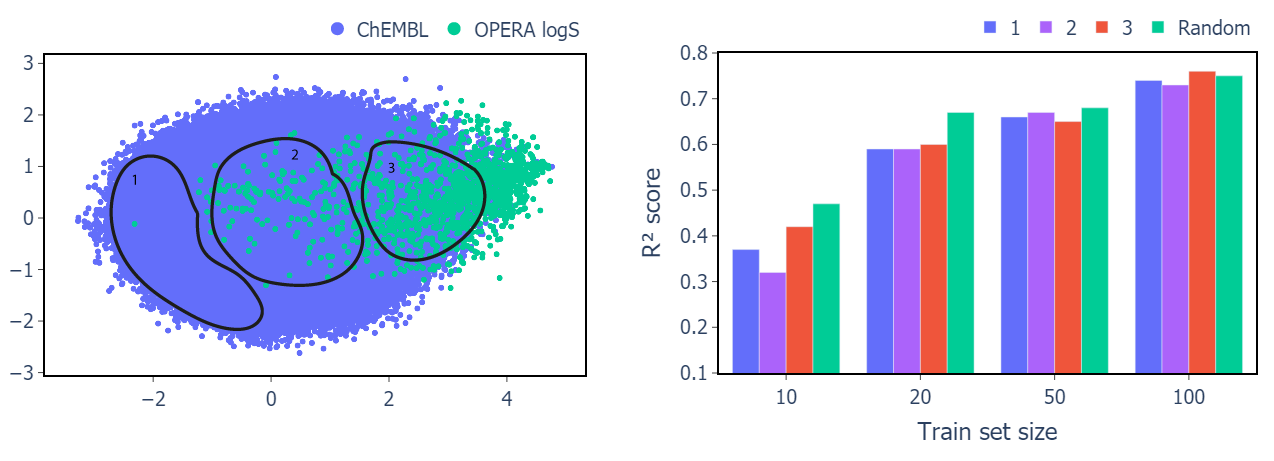}
  \caption{(a) PCA distribution of ChEMBL and OPERA molecules; (b) Acceptor models quality depending on donor composition.}
  \label{fig:fig4}
\end{figure}

We used the ranking numbers over five training iterations for each dataset size as the quality metric describing the stability and reproducibility of the solution. In this case, the splitting technique with the smallest ranking number should represent the most diverse property required to build a model with transfer learning. The best results were obtained with the highest variety of aromatic rings number in the training sets. However, the molecular weight and molecules conformation mobility (number of rotatable bonds) also proved to be important and even more significant for the smallest datasets. Also, it is noteworthy that datasets of 20 molecules were the most stable in relation to any selection criteria. That is, we can assume that 20 molecules with diverse sizes and number of flexible (rotatable bonds) and planar (aromatic ring) parts form a sufficient dataset for transfer learning. The next step to assess the impact of the donor dataset composition on the quality of the acceptor model was to understand the impact of the similarity between these datasets at a constant quantity. For this purpose, the ECFP4 values of the fingerprints were calculated for all molecules of the donor logP and acceptor logS datasets. The obtained vectors transformed into two-dimensional ones using principal components analysis (PCA) \cite{tipping61bishop} for demonstrative purposes. The results are presented in Fig. 4 (a). The three regions were numbered as the dataset intersection increased, from the area where there was almost no overlap (Area 1) to the area where the compositions of the datasets overlap significantly (Area 3). We extracted 10,000 molecules subsets from each of the regions, as well as a random one for the comparison. Using the obtained subsets, 4 donor models were obtained, which in turn were used to conduct transfer learning. The results are presented in Fig. 4 (b). Despite the presence of a weak correlation of R2 score with the structural similarity of the datasets, we can conclude that the similarity influence was negligible comparing to the influence of donor or acceptor datasets sizes.

Finally, besides the characteristics and size of the datasets, the similarity of donor/acceptor properties remains an important question to determine a possibility of applying the transfer learning approach. We  trained three groups of models: ‘pure’ GCNN models, without the use of transfer learning; the same models, transferred from the respective donor models; and models using a widespread combination of random forest algorithm \cite{breiman2001random, pedregosa2011scikit} and ECFP6 fingerprints \cite{extended2010fingerprints} for molecules feature extraction as a baseline. In Fig. 5 the regression and classification metrics comparison for datasets with best quality models among all splitting properties are presented.

\begin{figure}[ht]
  \centering
  \includegraphics[height=12cm]{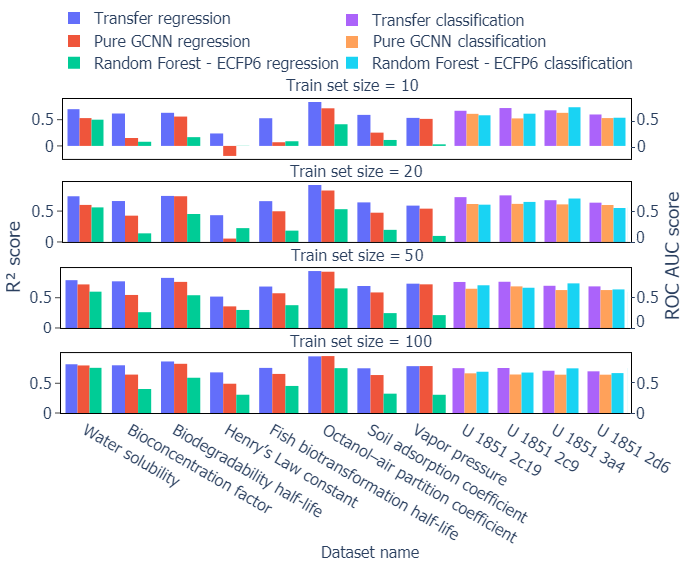}
  \caption{Evaluation metrics for models trained for acceptor properties with number of aromatic rings as acceptor dataset splitting property.}
  \label{fig:fig5}
\end{figure}

In vast majority of cases the efficiency of transfer model is superior to the other types, that proves that using the transfer learning methodology does improve the quality of the models of physicochemical properties as well as of biochemical ones. The results also clearly show the difference in the contribution of the transfer to the quality of the model, depending on the similarity of the donor and accentor properties. For classification problems, the contribution of transfer is practically the same, since all datasets belong to a similar biomedical problem. A similar situation for the OPERA datasets, transfer improvement strongly depends on the similarity of properties. For example, water solubility, bioconcentration factor, Henry’s law and octanol-air partition coefficient show the most efficiency from the transfer, since the acceptor properties are very similar to logP. At the same time quality of the transfer models for the vapor pressure is nearly the same as for pure GCNN, since the donor and acceptor properties are very different.

The difference in performance decreases with the increase of the size of the donor training set for the majority of cases, which corresponds to the general assumptions. It is also worth noting that some transfer models with smallest training set of 10 molecules show near zero quality. At the same time, with an increase in the size of the training set to 20, the models for the same properties showed much better results, which correlated with the conclusions obtained after analyzing the results for the sum of places from Figure 2 (b). It is important to note that the graph contains examples of transfer models, metrics of which exceed the metrics of other algorithms by at least 0.3, which is enough to make a high-quality model out of a mediocre one. There are also some properties (Bioconcentration factor, Fish biotransformation half-life) for which the results of non-transfer models are near zero when using the transfer technology go to 0.6 or more, i.e. to models of this quality that can already be used for coarse primary screenings. The OPERA datasets, which are not displayed in figure, are predominantly related to biological endpoints, so the use of logP as a donor did not lead to a significant improvement in the quality of the models and their performance is comparable to that of pure GCNN models.

Besides the quantitative metrics of the models’ performance, we also need to know their AD for practical usage. The AD itself strongly depends on the training data. The smaller and less diverse the training dataset, the smaller the AD of a resultant model. This makes it especially interesting to consider this characteristic for the transfer model, which trains on several dozens of objects. In this work, we applied the method for determining the AD described in \cite{sahigara2013defining}. First part of this method consists of determining the average distance between K neighbors (in our case K = 5) for every molecule in the train dataset – Dtrain. After this, for every molecule in test datasets we calculate average distance to the same K nearest neighbors in train datasets ($D_N$) and if $D_N$ < $D_{train}$, N-th molecule included in models AD. 

As one can see, transfer’s impact varies from 20\% to 35\% of observed molecules which could be considered as a significant improvement by the transfer modeling (Fig. 6). At the same time, it should be noted that the metrics of the models show more significant improvements in quality (Fig. 5), as well as the fact that correctly predicted molecules, which were not included in any AD, were not taken into account. Thus, we can conclude, that the results showed notable broadening of the model’s AD, though the ‘greedy’ nature of the AD calculation method indicates the need for further study of the quantitative effect of transfer learning on acceptor model’s applicability domain.

\begin{figure}[ht]
  \centering
  \includegraphics[height=6cm]{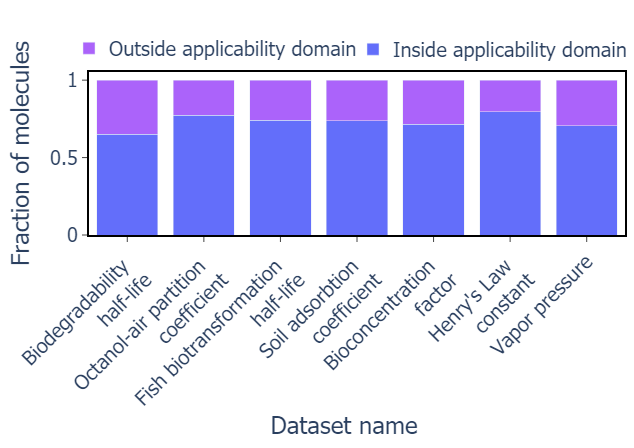}
  \caption{Estimated impact of donor model’s applicability domain on acceptor model’s performance.}
  \label{fig:fig6}
\end{figure}

\section{Conclusions}
We have demonstrated the performance of the transfer learning approach for modelling a variety of chemical endpoints from the water solubility of small organic molecules to the activities versus biological targets. In most cases, the proposed approach proved an ability to build workable models even on twenty molecules in the training set, where the use of more classical ML methods is not possible. The applicability domain of the trained models is also significantly broadened over the one, defined by the training data, that increases the suitability of the approach. 

Considering the number of tasks in which the amount of measured data points does not exceed one hundred, it is possible to say that transfer learning can be widely used in chemical applications of machine learning.

\section{Acknowledgments}
The research was carried out using the equipment of the shared research facilities of HPC computing resources at Lomonosov Moscow State University.

\bibliographystyle{unsrt}
\bibliography{references}

\end{document}